\pdfoutput=1

\documentclass[11pt]{article}

\usepackage[]{emnlp2021}

\usepackage{times}
\usepackage{latexsym}

\usepackage{xcolor}
\usepackage[colorinlistoftodos]{todonotes}

\usepackage[T1]{fontenc}

\usepackage[utf8]{inputenc}
\usepackage{booktabs}
\usepackage{graphics}
\usepackage{graphicx}
\usepackage{arydshln}
\usepackage{hyperref}
\usepackage{xcolor}
\usepackage{footnote}
\usepackage{footmisc}

\newcommand{\method}[0]{{IGA}}

\title{\method : An Intent-Guided Authoring Assistant}

\author{
    Simeng Sun$^1$\thanks{$^*$Most of the work done during an internship at Adobe.}
 \hspace{4mm} Wenlong Zhao$^1$ \hspace{4mm} Varun Manjunatha$^2$ \hspace{4mm} Rajiv Jain$^2$ \hspace{4mm}  \\
    \textbf{Vlad Morariu}$^2$ \hspace{3mm} \textbf{Franck Dernoncourt}$^2$ \hspace{3mm} \textbf{Balaji Vasan Srinivasan}$^2$ \hspace{3mm} \textbf{Mohit Iyyer}$^1$ \\
    University of Massachusetts Amherst$^1$ \hspace{1cm} Adobe Research$^2$\\
    {\tt \{simengsun,wenlongzhao,miyyer\}@cs.umass.edu}\\
    {\tt \{vmanjuna, rajijain, morariu, dernonco, balsrini\}@adobe.com}
}

\begin{document}
\maketitle
\begin{abstract}
While large-scale pretrained language models have significantly improved writing assistance functionalities such as autocomplete, more complex and controllable writing assistants have yet to be explored. We leverage advances in language modeling to build an interactive writing assistant that generates and rephrases text according to fine-grained author specifications. Users provide input to our \textbf{I}ntent-\textbf{G}uided \textbf{A}ssistant (\method) in the form of text interspersed with tags that correspond to specific rhetorical directives (e.g., adding description or contrast, or rephrasing a particular sentence). We fine-tune a language model on a dataset heuristically-labeled with author intent, which allows \method\ to fill in these tags with generated text that users can subsequently edit to their liking. A series of automatic and crowdsourced evaluations confirm the quality of \method's generated outputs, while a small-scale user study demonstrates author preference for \method\ over baseline methods in a creative writing task. We release our dataset, code, and demo to spur further research into AI-assisted writing.

\end{abstract}

\newcommand\para{\textsc{PARA}}
\newcommand\bio{\textsc{BIO}}
\newcommand\cause{\textsc{CAUSE}}
\newcommand\effect{\textsc{EFFECT}}
\newcommand\comparison{\textsc{CNTRA}} 
\newcommand\descrip{\textsc{DESCP}}
\newcommand\idiom{\textsc{IDIOM}}
\newcommand\mask{\textsc{MASK}}
\newcommand\newsroom{\textsc{Newsroom}}
\newcommand\sep{\textbf{\textit{<sep>}}}
\newcommand\ans{\textbf{\textit{<answer>}}}
\newcommand\genbutton{\texttt{Generate}}
\newcommand\addbutton{\texttt{+}}
\newcommand\base{\textsc{BASE}}
\newcommand\ilm{\textsc{ILM}}
\newcommand\oursys{\textsc{IGA}}
\newcommand{\nlp}[0]{{\textsc{NLP }}}

\section{Introduction}

Writing can be both an exhilarating, creative experience as well as a frustrating slog. Can recent advances in neural language modeling help improve the human writing experience? Pretrained Transformer language models~\citep{radford2019language}  have improved  writing aids such as email ``autocomplete'' \cite{gmailsmartcompose}, while commercial tools such as Grammarly and Microsoft Editor can rewrite full sentences to increase clarity.\footnote{As these systems are not open-sourced, it is unclear how exactly they are implemented.} 

\definecolor{orange}{RGB}{254, 228, 205}
\definecolor{green}{RGB}{213, 232, 212}
\definecolor{yellow}{RGB}{255, 242, 204}
\definecolor{red}{RGB}{248, 206, 204}
\setlength{\fboxsep}{0pt}
\begin{figure}[t]
    \scalebox{0.47}{\includegraphics{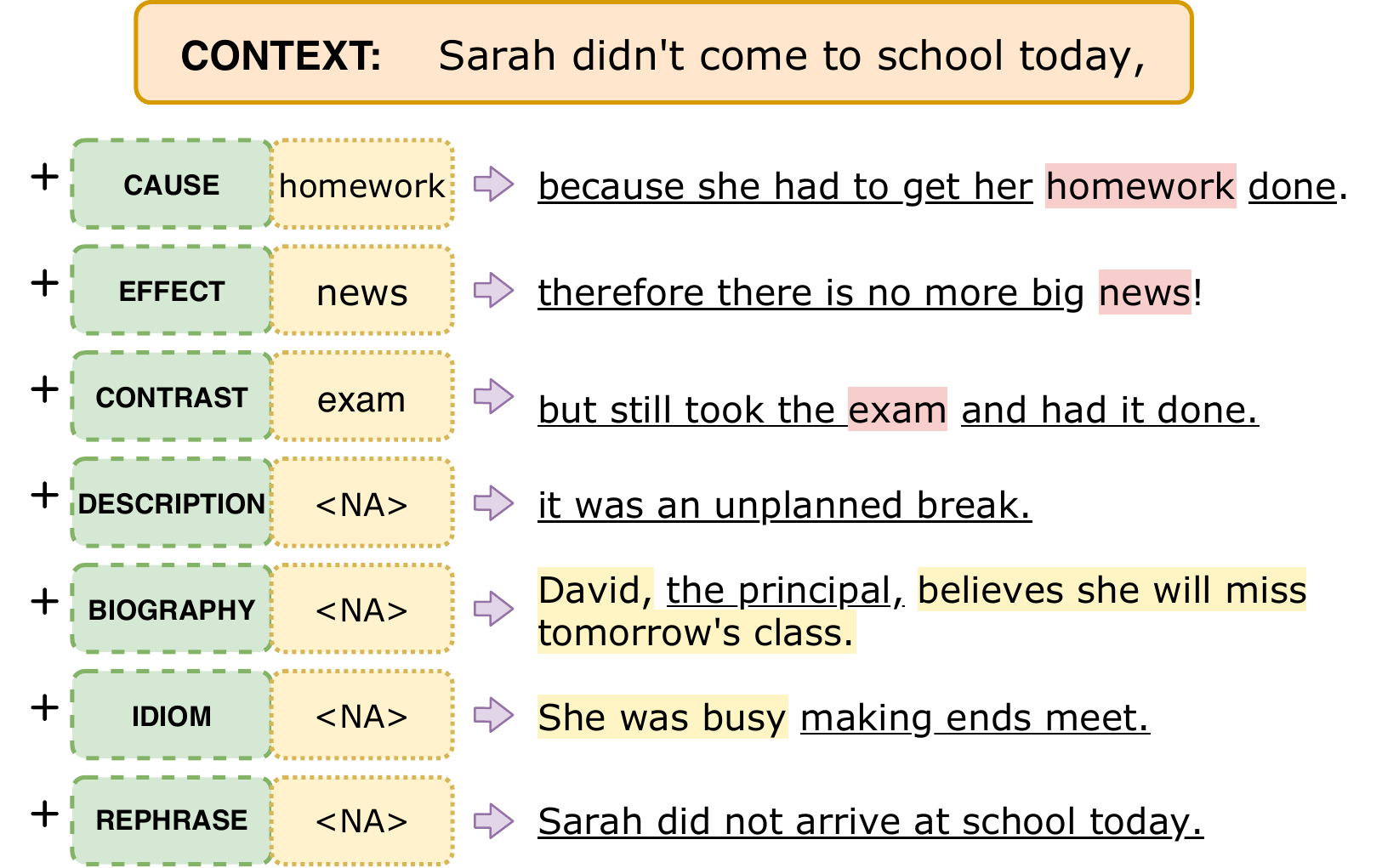}}
    \caption{General concept overview of our \textbf{I}ntent-\textbf{G}uided \textbf{A}uthoring assistant \oursys. Given \colorbox{orange}{\strut context}, by specifying different writing intents, the system generates \underline{output} satisfying the \colorbox{green}{\strut intent}. In addition to well-formed \colorbox{yellow}{\strut  sentence fragments}, \colorbox{red}{keywords} can also be part of user input, serving as arguments for the \colorbox{green}{\strut  intents}, and are preserved in the output.}
    \label{fig:figure1}
\end{figure}

Few existing writing assistants provide support for the underlying cognitive process of writing~\citep{greer2016an}. In this paper, we explore more advanced writing assistance functions: specifically, we build an authoring assistant capable of following fine-grained user directives (e.g., add descriptive text, use idiomatic language, or paraphrase a clunky bit of wording).  
Our system, the Intent-Guided Assistant (IGA), combines controllable text generation with text infilling~\cite{zhu2019text,keskarCTRL2019,lewis-etal-2020-bart,donahue-etal-2020-enabling}; more specifically, we adapt the tag-based control of ~\citet{keshar-2019-ctrl} to include a set of rhetorical directives that our model learns to infill with relevant and fluent text. Our system can handle the following author-guided tags: \emph{cause, effect, concession (contrast), description, biography, idiom}, and \emph{rephrase}. User input to \method\ can be as simple as a list of keywords and does not have to include well-formed text (Figure \ref{fig:figure1}).

We train \method\ in supervised fashion by creating a large multi-domain dataset in which spans corresponding to particular directives are replaced with a single tag: for the input \emph{``It was raining \underline{<description>} trees"}, the ground-truth completion could be ``\emph{It was raining \underline{, the} trees \underline{were swaying and the wind was oppressive}."} To build our dataset, we use heuristics based on lexical and syntactic choice to isolate spans corresponding to each directive. For the above example, we extract the first simple declarative clause, then highlight the span that contains words (such as ``oppressive") in a large list of adjectives, and finally extract keywords such as ``trees" using a keyword extractor. At inference time, our model can flexibly take any tag as input: given \emph{``It was raining \underline{<contrast>} trees"}, for example, our model inserts a contrastive clause to produce \emph{``It was raining \underline{but still the} trees \underline{were not wet}"}. 


To evaluate the effectiveness and usability of our AI-assisted writing paradigm, we design \method\ to be interactive, in the spirit of human-AI co-authoring. In addition to automatic and crowdsourced evaluations that demonstrate \method's output quality, we perform a user study in which participants make use of our system for creative storytelling (Section \ref{sec:human-eval}). Our results show most users  prefer writing with assistance from \method\ compared to writing from scratch or with a non-controllable infilling model.

Our contributions are as follows: 
\begin{enumerate}
\itemsep0em 
\item We design \method, an authoring assistant capable of controlled text generation based on explicit rhetorical directives  specified by the author.
\item To train \method, we create a large dataset (75M tokens) of text heuristically-labeled with author intent, sourced from multiple repositories. This dataset is made publicly available to facilitate future research on AI-enabled author assistance.\footnote{Dataset and models can be found at \url{https://github.com/SimengSun/IGA-An-Intent-Guided-Authoring-Assistant}}
\item We validate the usefulness of \method\ through automatic and crowdsourced evaluations as well as a user study involving creative writing. 
\end{enumerate}

\begin{figure*}
    \centering
    \includegraphics[width=\textwidth]{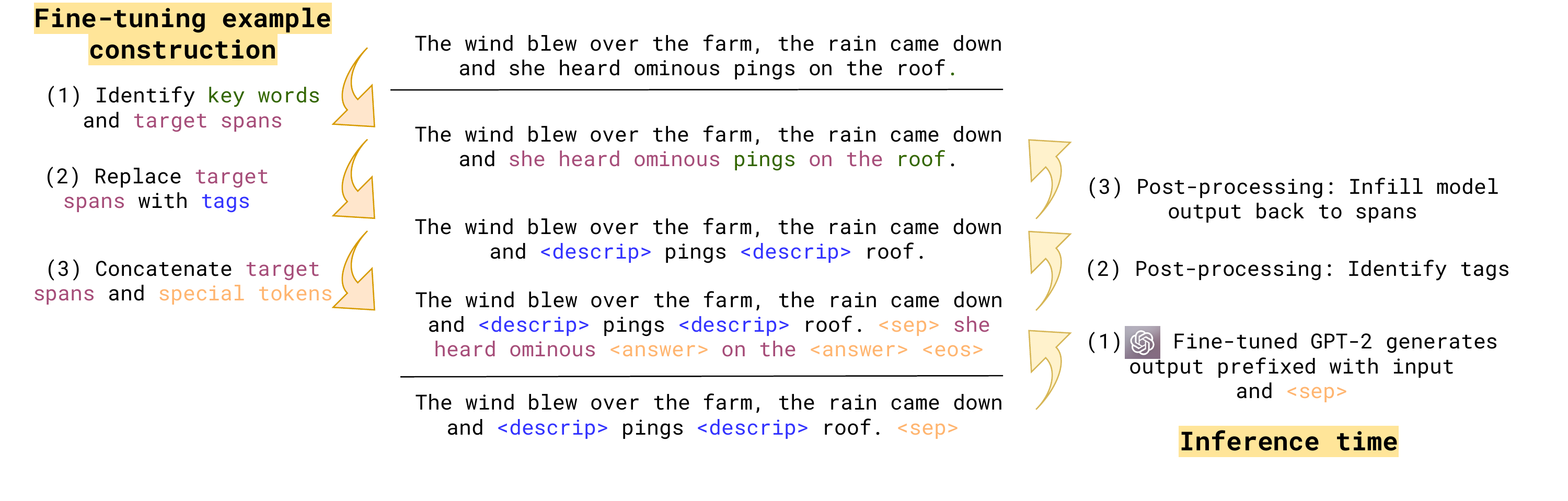}
    \caption{On the left, we show how each example is constructed for fine-tuning. 
    On the right, we show how the final output is constructed by post-processing the output of a fine-tuned GPT-2 model at inference time.
    }
    \label{fig:template-construct}
\end{figure*}



\section{Related Work}
\label{sec:related}

\subsection{Theories of writing} 
Numerous studies within the humanities focus on modeling the process of effective writing \cite{flower-hayes, prewriting, grabe-kaplan}.
We base the design of \method\ on the widely cited and reproducible ``cognitive process theory of writing'' of~\citet{flower-hayes}, which was made more comprehensive in the review work by~\citet{Becker2006ARO}. This theory posits that writing is a non-linear process that consists broadly of three steps : \emph{planning}, \emph{translating} and \emph{reviewing}. The \emph{planning} phase involves accessing one’s knowledge of the topic and target audience to formulate a rough outline of the eventual output. The actual rendering of the text on paper is called \emph{translating}, while the \emph{reviewing} phase consists of making edits or revisions to the output. All of these steps happen concurrently, under the influence of a \emph{monitor}.  In our work, a human author and a language model jointly participate in  the planning and translating phase, while the human (by means of an editable output interface) reviews and monitors the process. 

 
\subsection{Infilling language models} 
The actual implementation of \method\ relies on controllable text \emph{infilling} via language modeling. 
The ability of large-scale language models to generate fluent and coherent text has been demonstrated in several prior works~\cite{radford2019language, brown2020language, zellers2019neuralfakenews} when given only a few words or a sentence as a prompt. More recent research has addressed the inability of these models to infill text, or insert new words/tokens between tokens that already exist~\cite{donahue-etal-2020-enabling, zhu2019text, huang-etal-2020-inset, pmlr-v97-stern19a, pmlr-v97-welleck19a, zhang-etal-2020-pointer, liao-etal-2020-probabilistically, moryossef-etal-2019-step}. 
In this vein,~\citet{rashkin-etal-2020-plotmachines} generate coherent stories given just bullet-point plot outlines, while \citet{cai-etal-2019-skeleton} perform token insertion using a retrieval engine in combination with a language model for dialogue agents. Unlike \method, however, none of this prior work can control generation using high-level rhetorical directives specified by an author.

\subsection{Controllable Text Generation}

\method\ conditions its generated text on tags, which has previously been done for left-to-right language models. For example,~\citet{Dathathri2020Plug} combine a large language model with an attribute discriminator to generate text that obeys certain sentiments or topics.
Meanwhile, expanding the \emph{control codes} proposed in \citet{keshar-2019-ctrl}, ~\citet{Krause2020GeDiGD} develop a model that can generalize to new control codes,
while the Megatron-CNTRL model~\citep{Megatron} control the output with predicted keyword.
In contrast to these works, \method\ focuses on fine-grained, intra-sentential controlled infilling.  Previous work has also explored controlling stylistic parameters~\cite{ficler-goldberg-2017-controlling} and syntactic structures~\cite{iyyer-etal-2018-adversarial,goyal-durrett-2020-neural}.

\section{Intent-Guided Assistant}

\method\ extends text infilling models with fine-grained rhetorical control. Specifically, we build on the Infilling Language Model (ILM) of~\citet{donahue-etal-2020-enabling}, which fine-tunes an off-the-shelf language model such as GPT-2 on a dataset of text with masked spans. 
To continue with our running example, 
the input to this model is the sequence \emph{``It was raining} \textit{\textbf{<mask>}} \emph{trees"},
and a target output is \emph{``and getting harder to see the"}.
At inference time, the blanks are substituted with the words predicted by the LM and combined with the input in a post-processing step to generate the final output: \emph{``It was raining \underline{and getting harder to see the} trees"}. 

Building on this framework, we fine-tune an off-the-shelf GPT-2 medium model\footnote{This model has 431M parameters.} on our dataset (described in Section~\ref{sec:dataset})  created for generating text conditioned on author intent. Instead of replacing spans with a generic \textit{\textbf{<mask>}} token as in ILM, we replace spans with more fine-grained tags corresponding to rhetorical directives. For fine-tuning, we concatenate a tag-replaced sentence with the ground-truth spans that should be infilled using a special separator token \sep, as in ~\citet{donahue-etal-2020-enabling}. 
If multiple tags are used in the input, the multiple ground-truth spans following the \sep\ token are separated by special \ans\ tokens. At inference time, the model is fed a tag-replaced sentence and the \sep\ token, from which it generates  span(s) to infill the input tags. In a post-processing step, we replace the tags with the generated answer spans. 
Figure \ref{fig:template-construct} summarizes the fine-tuning process (left) and the inference process (right). During inference time, we use top-$k$ sampling~\cite{fan-etal-2018-hierarchical} with $k$ fixed to 40.

To sum up, unlike the standard GPT-2 model which only supports strict left-to-right generation, both ILM and IGA are capable of text-infilling. Moreover, IGA has finer control over author writing intents than ILM, which further narrows down the output generation space. Although providing GPT-2 and ILM  with specific discourse markers may result in output conforming to certain intents, they are less flexible than IGA, which can sample from multiple discourse cues.

\section{Dataset} \label{sec:dataset}

\begin{table}[ht!]
    \centering
    \scalebox{0.7}{
    \begin{tabular}{ccccccccc}
    \toprule
          & \para & \bio & \cause & \effect  \\ \midrule
        Train & 148,621 & 200,000 & 200,000 & 108,328  \\
        Dev & 500 & 500 & 500 & 500  \\
        Test & 500 & 500 & 500 & 500 \\ \midrule
        & \comparison & \descrip & \idiom & \textbf{Total} \\ \midrule
        Train & 200,000 & 198,760 & 176,738 & 1,232,447\\
        Dev & 500 & 500 & 500 & 3500\\
        Test  & 500 & 500 & 500 & 3500\\
        \bottomrule
    \end{tabular}
    }
    \caption{Number of examples per tag and in total.}
    \label{tab:data_stats}
\end{table}

\begin{table*}[ht!]
\centering
    \scalebox{0.9}{
    \begin{tabular}{p{0.1\linewidth} p{0.9\linewidth}}
    \toprule
       Tag  &  System input and output \\ \midrule
       \para  & \textbf{Input:} \textcolor{blue}{<paraphrase>} The pandemic has caused very serious problems. \textcolor{blue}{<paraphrase>} \\
       & \textbf{Output:} \textcolor{blue}{The pandemic has brought severe economic, social and political effects that have seriously affected many countries.}\\
        \hdashline       
       \bio  & \textbf{Input:} Oria, \textcolor{blue}{<biography>} , mentions that technically only humans can cry in response to emotional state. \\
       & \textbf{Output:} Oria, \textcolor{blue}{a psychologist specializing in the study of human emotion}, mentions that technically only humans can cry in response to emotional state.\\
       \hdashline
       \cause  &  \textbf{Input:} This is a really good book \textcolor{blue}{<cause>} plot \textcolor{blue}{<cause>} .\\
       & \textbf{Output:} This is a really good book \textcolor{blue}{because the} plot \textcolor{blue}{is always so well written and never predictable}.\\
       
       \hdashline
       \effect  & \textbf{Input:} Janet had suffered a traumatic brain injury in a car accident in 1988, \textcolor{blue}{<effect>}. \\
       & \textbf{Output:} Janet had suffered a traumatic brain injury in a car accident in 1988, \textcolor{blue}{thus had no memory of who she was or what happened}.\\
       \hdashline
       
       \comparison  & \textbf{Input:} The castle was built in 1865, \textcolor{blue}{<contrast> }.\\
       & \textbf{Output:} The castle was built in 1865, \textcolor{blue}{though a full-scale modern restoration has been underway for the past four years.}\\
       
       \hdashline
       
       \descrip  &  \textbf{Input:} \textcolor{blue}{<description>} individual \textcolor{blue}{<description>} beliefs \textcolor{blue}{<description>} wrong .\\
       & \textbf{Output:} \textcolor{blue}{There are} individual \textcolor{blue}{and social} beliefs \textcolor{blue}{that should lead us to be skeptical of the facts and the} wrong.\\
       
       \hdashline
       
       \idiom  &  \textbf{Input:} This report only shows the \textcolor{blue}{<idiom>} , as many details can only be uncovered if you sign the document.\\
       & \textbf{Output:} This report only shows the \textcolor{blue}{tip of the iceberg} , as many details can only be uncovered if you sign the document.\\
    \bottomrule
    \end{tabular}
    }
    \caption{Example output of each tag from {\oursys}.}
    \label{tab:example_output}
\end{table*}

The novelty of \method\ lies not in its model architecture but the way in which we supervise it to enable controlled  fine-grained text infilling. 
We construct the fine-tuning dataset primarily by heuristically mining the \newsroom\ corpus ~\cite{grusky-etal-2018-newsroom}, the largest available summarization dataset with 1.3 million news articles. We also collect partial data from ParaNMT-50M~\cite{wieting-gimpel-2018-paranmt}, WikiLarge~\cite{zhang-lapata-2017-sentence} for ``sentence embellishment'' writing intent, and PoMo~\cite{kang-etal-2019-pomo} to extract post-modifier that comes after an entity. 
Our dataset (statistics shown in Table \ref{tab:data_stats}) contains 75M tokens with a mean example length of 60.5 words tokenized with NLTK~\cite{BirdKleinLoper09}. 

\paragraph{Choosing a collection of tags:} 
Before we start collecting data, we conduct an internal survey with potential users of our system to determine what writing assistance functions they would most benefit from. We surveyed nine NLP researchers about their opinions on the ideal functionality of an authoring assistant. 
After removing simple functions such as generating synonyms, antonyms, adjectives, and adverbs, which are already implemented in existing writing assistant tools,
we condense the most requested writing intents into seven tags (described in detail below; examples provided in Table \ref{tab:example_output}). Only one of them (\para) is heavily constrained by semantic content, and the rest involve open-ended generation loosely constrained by keywords and intent.

\subsection{Data collection for each writing intent}

\paragraph{\cause:} This tag helps an author invent a reason for the occurrence of an event. Clauses with \cause\ intent usually follow words/phrases like `because' or `due to'. We manually extracted 16 markers, many from the discourse marker list in~\citet{sileo-etal-2019-mining}, and then mine \newsroom\ ~\cite{grusky-etal-2018-newsroom} to find sentences that match any of the markers. For all mined examples, we also preserve the previous sentence as the context of the matched sentence. Simple declarative clauses that start with matched discourse markers are extracted through shift-reduce constituency parser ZPar~\cite{zhang-clark-2011-syntactic}. The YAKE algorithm is later applied to those clauses for keyword extraction. 

\paragraph{\effect:} As a conjugate writing intent of \cause, \effect\ is used when one needs to describe the result or consequence. It usually co-occurs with the words/phrases such as `therefore' and `as a result'. Similar to \cause, we manually select 15 discourse markers that signify \effect
, mine sentences from \newsroom, and highlight spans of interest using the same process as before. Specifically, we extract declarative clauses based on the constituent labels returned from the parser. Then, we decide the intent according to the starting markers of those clauses. Inside each clause, we run YAKE, an unsupervised keyword extraction algorithm, to extract keywords for the clause. All markers are included in Appendix \ref{appendix:discourse-markers}.

\paragraph{\comparison:} Comparison is a commonly used writing technique that encompasses two separate intents: concession and contrast. Concession refers to the unexpectedness of a consequence~\cite{webber2019penn}: words/phrases such as ``although'' and ``even though'' raise an expectation curbed by the rest of the sentence. Contrast is often confused with concession, but its markers include ``by comparison'', ``in contrast'', etc. We manually select 31 concession markers and six comparison markers, and mine the Newsroom corpus with these to obtain labeled data. 

\paragraph{\descrip:} Descriptive details are important for creative writing and can help embellish written output. To curate data for this tag, we first collect 27K descriptive adjectives based on morphological rules.\footnote{Adjectives are extracted from English word frequency list \url{https://norvig.com/ngrams/count_1w.txt}} We then mine sentences from \newsroom\ and extract simple declarative clauses that contain the matched adjective(s). The spans are filtered to be greater than five tokens.

\paragraph{\idiom:} Idiomatic language makes writing more vivid and imaginative. We collect 3000 idioms online\footnote{Idioms are extracted from \url{https://7esl.com/english-idioms/} and \url{https://www.phrases.org.uk/meanings/phrases-and-sayings-list.html}} and mine sentences from \newsroom\ that match any of the idioms. In order to include variants of the raw idiom, e.g. ``apple of \textit{one's} eye'', we apply regular expressions to match various possessive forms. 

\paragraph{\bio:} Biographical post-modifiers are commonly used to provide a brief summary of a previously mentioned named entity. For example, ``co-founder of Microsoft Corporation'' fills in the blank span of the sentence ``Bill Gates, \underline{\hspace{1cm}} , has pursued a number of philanthropic endeavors''. PoMo~\cite{kang-etal-2019-pomo} is an existing dataset that aligns with this writing intent. It contains sentences with post-modifiers and facts about named entities extracted from Wikidata. We use the textual data in PoMo, replace the post-modifier with \textbf{\textit{<biography>}} and treat the ground-truth post-modifier as target span. 

\paragraph{\para:} Paraphrasing is a common method by which authors improve their draft writing \cite{flower-hayes, artful-tufte}. Unlike sentence simplification, the intent of our \textit{<paraphrase>} tag is to paraphrase with \emph{improved} writing quality, similar to embellishment. We construct parallel data for this tag by combining ParaNMT-50M~\citep{wieting-gimpel-2018-paranmt}, a large corpus consisting of back-translated sentence pairs, with WikiLarge, a sentence simplification dataset with parallel simple and complex sentences. The original sentence in ParaNMT-50M and complex sentence in WikiLarge are treated as targets, while the back-translated sentence and the simplified sentence are used as the source. We use BLEURT~\cite{sellam-etal-2020-bleurt} to filter noisy pairs from ParaNMT-50M,\footnote{We set BLEURT threshold to be (0.7, 0.9) to avoid semantically dissimilar sentences and sentences without too much change.}
discarding pairs whose word-level edit distance is less than five. To further encourage  complex paraphrases, we require the reference sentence to have more low-frequency words than the candidate sentence.

\section{Evaluation against references}

As an initial comparison of \method\ and \ilm, we evaluate the generated outputs of each model against reference completions from our dataset, both automatically and through a crowdsourced evaluation. We acknowledge that this type of evaluation (especially using automatic metrics) is limited for open-ended generation tasks like ours~\cite{fan-etal-2018-hierarchical,akoury-etal-2020-storium,rashkin-etal-2020-plotmachines}, which is why we also conduct an in-depth user study in Section~\ref{sec:human-eval}. While results of these evaluations cannot reflect how practical \oursys\ can be used as an authoring assistant, they do indicate that \oursys\ is more constrained than \ilm\ and produces output that better fulfills the writing intents. 


\begin{table}[]
    \centering
    \scalebox{0.8}{
    \begin{tabular}{lcccccc}
    \toprule
\multicolumn{1}{l}{} & \multicolumn{2}{c}{ROUGE-2} & \multicolumn{2}{c}{BLEU-2}  & \multicolumn{2}{c}{Length}\\\cmidrule(lr){2-3} \cmidrule(lr){4-5} \cmidrule(lr){6-7}
\multicolumn{1}{l}{} & \ilm          & \oursys           & \ilm          & \oursys      & \ilm          & \oursys     \\\midrule
\bio                 & \textbf{10.4}        & 9.9         & \textbf{47.7}        & 44.2    & 6.3 & 6.0  \\
\cause               & 4.1         & \textbf{9.0}         & 35.0        & \textbf{37.1}    & 10.1 & 10.2   \\
\effect               & 5.2         &\textbf{6.6}        & 37.2        & \textbf{37.8}      & 13.2 & 13.4 \\
\comparison             & 4.2         & \textbf{4.9}         & 32.3        & \textbf{34.6}   & 10.3 & 10.3    \\
\descrip              & 2.1         & \textbf{2.2}        & 23.4        & \textbf{23.6}    & 8.9 & 8.9   \\
\idiom                & 33.7        & \textbf{37.8}        & 62.3        & \textbf{64.5}  & 3.0 & 2.7 \\\bottomrule
\end{tabular}
    }
    \caption{ROUGE-2, self-BLEU2, and total number of infilled tokens of each example on test set.}
    \label{tab:auto-res}
\end{table}

\subsection{Automatic evaluation}


We compare \oursys\ with \ilm\ on automatic metrics such as ROUGE~\cite{lin-2004-rouge} and self-BLEU~\cite{ self-bleu} following~\citet{rashkin-etal-2020-plotmachines}, computing both scores against reference completions from the validation set.\footnote{All automatic metrics are computed only on the infilled spans, excluding the context.} On all but the BIO tag, \oursys\ achieves higher ROUGE and self-BLEU scores than \ilm\ (Table \ref{tab:auto-res}), which shows that \oursys\ outputs have higher coverage and lower diversity, respectively, without differing considerably in length. This result indicates that \oursys\ is indeed conditioning its output on the tags to produce more constrained outputs. 
Since the infilled spans of \bio\ are strictly post-modifiers that follow a very specific structure (i.e., enclosed by two commas), the superior performance of \ilm\ indicates that it memorizes this simple form of construction without requiring a separate tag input.

\para\ is the only substitution-based tag in our system and is not supported by \ilm. Therefore, we compare performance of \para\ with the state-of-the-art paraphraser \textsc{STRAP} released by~\citet{krishna-etal-2020-reformulating} with the default nucleus sampling $p=0.6$. We compute BLEURT scores to check semantic similarity, as well as BLEU~\cite{Papineni-etal-2002-bleu}, self-BLEU~\cite{sun-zhou-2012-joint}, and iBLEU~\cite{sun-zhou-2012-joint} with $\alpha=0.8$ to check the diversity of output. Table \ref{tab:para-auto-eval} indicates that \oursys\ outperforms \textsc{STRAP} in all dimensions. We hypothesize that this is primarily because 
the diverse paraphraser in \textsc{STRAP} normalizes (and often simplifies) stylized text, while our \para\ tag is associated with complex, embellished paraphrases during fine-tuning.

\begin{table}[]
    \centering
    \scalebox{0.85}{
        \begin{tabular}{ccccc}
       \toprule
          & BLEURT & BLEU & s-BLEU & i-BLEU\\\midrule
\textsc{STRAP} & 0.31 & 21.79 & 19.50 & 9.4\\
\oursys\ & \textbf{0.53} & \textbf{33.90} & \textbf{13.97} & \textbf{15.46}\\
    \bottomrule
    \end{tabular}
    }
    \caption{BLEURT, BLEU, self-BLEU, iBLEU scores comparison of \textsc{strap}~\cite{krishna-etal-2020-reformulating} and \oursys\ on \para\ validation set. }
    \label{tab:para-auto-eval}
\end{table}

\subsection{Intrinsic crowdsourced evaluation}
The above automatic evaluations can only tell us so much about \method's capabilities. Many of our tags (e.g., \descrip, \cause) are open-ended, which results in a large space of acceptable outputs. Thus, measuring similarity to ground-truth span completions is not as suitable for our task as it is for more constrained tasks such as machine translation or summarization. 
In this section, we shift to human evaluation as a way to learn more about the behavior and usefulness of \method. 

We begin with a small-scale intrinsic evaluation to get a sense of the generation quality and adequacy of an output in fulfilling a writing intent.
We randomly select 50 examples from the test set of each tag and generate outputs using both \ilm\ and \oursys\ for each example. For each output, three Mechanical Turkers are shown the gold completion as well as the generated text and asked to choose which is more fluent, coherent, and adequate at fulfilling the author intent, using a five-point scale to measure fine-grained preference.\footnote{We employ workers with an approval rate higher than 96\% and total approved HITS greater than 1000. Each rater is rewarded for \$0.1 per HIT.}
The results of this task are inconclusive:
although \oursys\ outputs for most tags are more often preferred than those from \ilm\ across these dimensions (see Appendix \ref{appendix:mturk} for specifics), especially in terms of adequacy, the subjective nature of the task yields low agreement among annotators.\footnote{The best Fleiss $\kappa$~\cite{fleiss1971measuring}  across all tags was only slightly over 0.2, which indicates slight agreement~\cite{landis1977measurement}.} In general, annotators found the task difficult and most often chose to express no preference. 

\section{User study} \label{sec:human-eval}

\begin{figure*}[ht!]
    \centering
        \includegraphics[width=\textwidth]{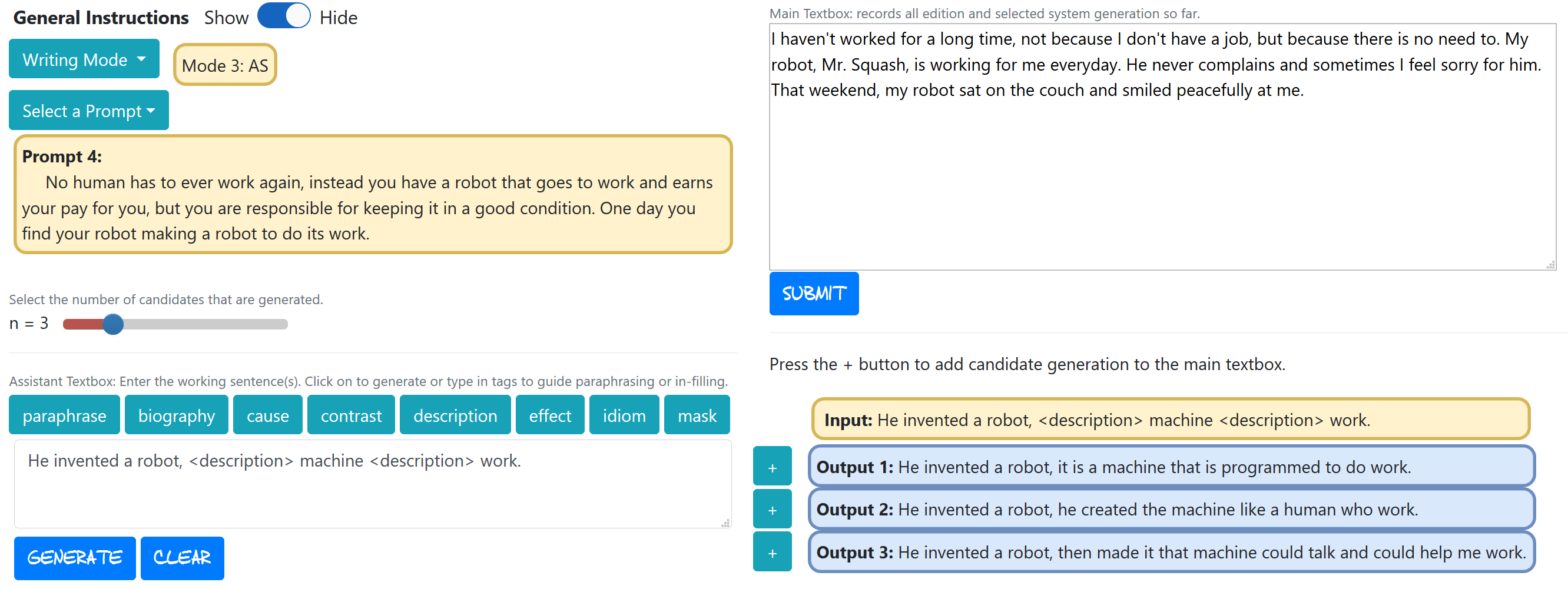}
    \caption{User interface screenshot. The bottom-left Assistant textbox accepts intent marked-up sentences. Model output can be added to the top-right Main textbox through the `+' button. All edits and sentences without machine assistance happen in the Main textbox.}
    \label{fig:sys-overview}
\end{figure*}
Due to the limitations of the previous evaluations, we launch a user study in the same spirit as~\citet{10.1145/3172944.3172983} to understand the interactive behavior of real users. We measure whether human authors benefit from AI-assisted writing, and whether they prefer intent-guided generation to the uncontrolled ILM model. We design an interactive web demo that allows participants to write with the help of each model, logging their behavior (e.g., queries to the model, edits made on generated text) and self-reported feedback.

Our interactive demo is inspired by markup language editors such as Overleaf\footnote{\url{https://www.overleaf.com/}} or Lyx\footnote{\url{https://www.lyx.org/}}; a screenshot of the interface is shown in Figure~\ref{fig:sys-overview}. In the textbox to the left, users type sentences with any of our supported tags. After clicking \genbutton, the model's output will be displayed on the right hand side\footnote{It takes $\sim$1s to display three model outputs.}. By default, three samples are shown to the user, although they can increase the number of samples if they wish. After a user selects a sample, it is appended to the existing text in the top-right textbox, which contains all of the text the user has already written. Users can then edit the sample (or completely delete it) and write continuations directly into the top-right textbox. This process repeats every time the user decides to use a tag to obtain model-generated text. On the backend, the input fed to our \method\ model is the concatenation of content in the main textbox (i.e., context) and the input in the assistant box, truncated at 300 tokens.

\subsection{User study design}

We recruited twelve computer science graduate students for our user study, seven of whom are native English speakers.\footnote{Each participant is rewarded with a \$30 Amazon gift card.} Nine of the twelve participants in the user study had creative writing experience in English prior to the evaluation, three participants had taken creative writing classes, and one was trained in media writing. We asked each participant to write short stories in response to prompts selected from WritingPrompts~\cite{fan-etal-2018-hierarchical}, a large dataset of stories written by users on Reddit. This task is suitable for our user study because creative writing requires diverse rhetorical directives while also not placing as much of an emphasis on world knowledge on the part of the participant (unlike writing a news article, for instance).

We ask each participant to write responses to three different prompts, where for each prompt they use one of three different writing modes: 
(1) \textbf{\base:} writing from scratch without any AI assistance,
(2) \textbf{\ilm:} writing assisted only with the \textbf{\textit{<mask>}} tag, and
(3) \textbf{\oursys:} writing assisted with multiple tags. To study how often users use intent-guided generation instead of uncontrolled generation when given a choice, we also include the \textbf{\textit{<mask>}} tag in the \oursys\ mode by simply producing outputs from the \ilm\ model. We randomize the order of modes across subjects to mitigate respondent fatigue~\citep{lavrakas2008encyclopedia} (e.g., one participant may write their responses using \textbf{BASE, ILM, IGA} while another might use \textbf{IGA, BASE, ILM}).

Each evaluation session lasts for approximately one hour. Before each session, the participant is instructed to read a tutorial document, which describes the system's layout and the usage of each tag. During each evaluation session, they first go over an interactive tutorial to experiment with each tag, either with provided examples or examples they invent themselves, and then start the main writing tasks. The purpose of the tutorial is to thoroughly familiarize participants with each model so they do not have to learn on the fly.

During the AI-assisted writing phase, we do \emph{not} require participants to write every sentence with the tags, or even use the system at all if they choose not to (e.g., they can write their whole response from scratch). We do require them to write at least ten sentences in response to each prompt.
In each evaluation session, we record the following metrics to understand how the participants interact with the systems: 
\begin{enumerate}
\itemsep0em 
\item   \# of clicks on the \genbutton\ button, which takes the user-tagged sentence and outputs multiple (sampled) completions
\item  \# of clicks on the \addbutton\ button, which adds a sampled completion to the Main textbox
\item \# of sentences written without any assistance 
\item \#  of model-generated tokens that were kept and deleted by the author in the Main textbox
\item  \#  of novel tokens inserted by an author within a model-generated completion.
\end{enumerate}

\begin{figure}
    \centering\includegraphics[width=0.5\textwidth]{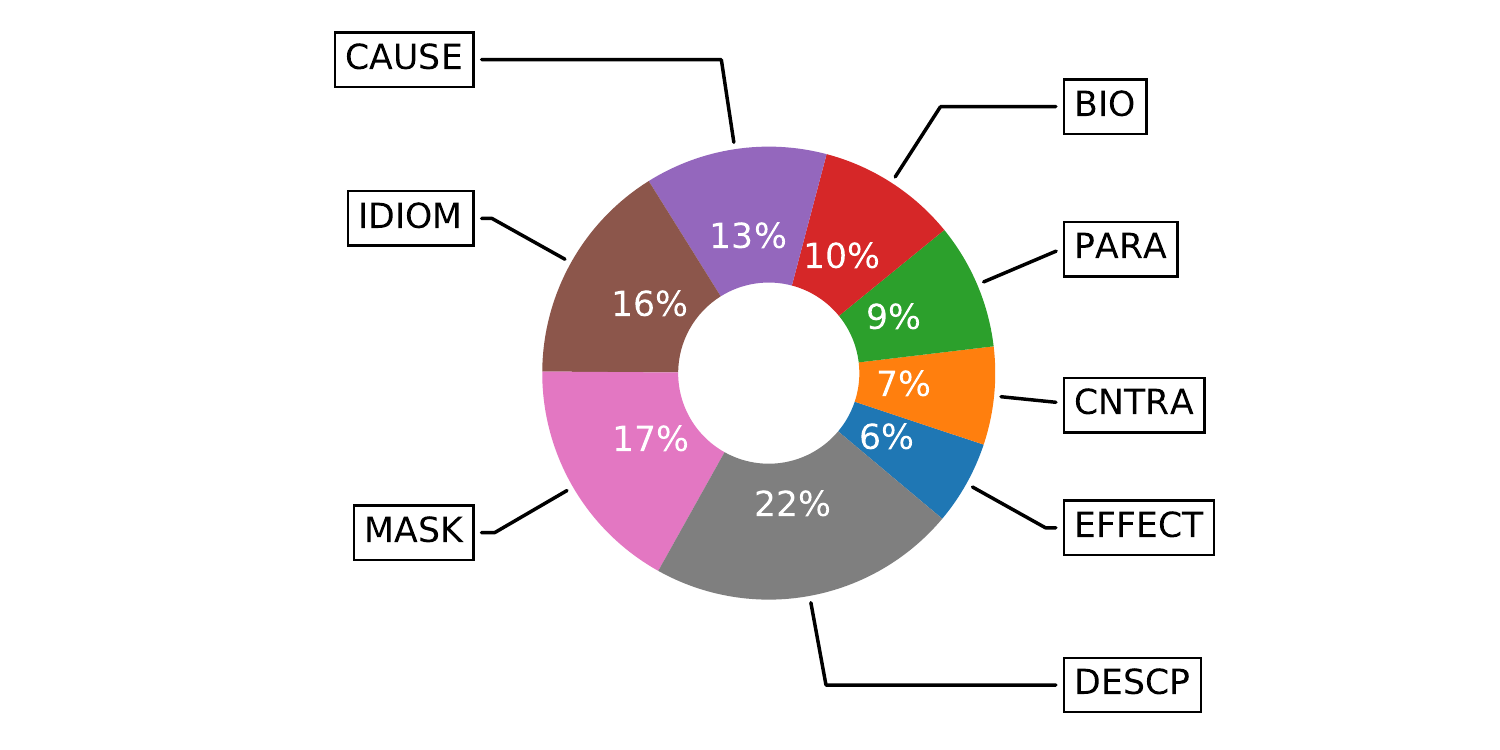}
    \caption{DESCP and the uncontrolled MASK were the most commonly used \method\ tags from our user study.}
    \label{fig:tag-pie-chart}
\end{figure}

We report the average number of tokens and sentences in the responses (Table~\ref{tab:user-study-stats}), the average number of clicks per sentence (Table~\ref{tab:user-study-click}), tag usage of all AI-assisted output (Figure \ref{fig:tag-pie-chart}), and unigram precision, recall, and F1-scores of each intent tag (Table \ref{tab:user-study-edit}). In summary, users interact more frequently with \oursys\ than \ilm, generating more content ($\sim3$ more sentences per session and $\sim30$ more tokens in each response), and more of their sentences on average are AI-assisted with \oursys\ (8.0 compared to 6.6). 
Additionally, \oursys\ generations are far less likely to be edited than those from \ilm: Table \ref{tab:user-study-edit} shows  69\% of the generated tokens are preserved in \ilm\ mode, compared to $\sim87\%$ in \oursys\ mode (averaged across all tags). Interestingly, when equipped with the intent-based tags in \oursys\ mode, the output of the uncontrolled  \textbf{\textit{<mask>}} tag, the second most often used tag in \oursys\ mode as shown in Figure \ref{fig:tag-pie-chart}, is more likely to be accepted by users than in the \ilm\ (80 vs. 68). This is likely because when the users use the \textbf{\textit{<mask>}} tag under \oursys, they indeed have no clear intents, and are thus more likely to accept intent-free generation.

\begin{table}[t]
    \centering
    \scalebox{0.85}{
    \begin{tabular}{lccccc}
    \toprule
      Mode  & \# Tokens & \# Sents & \# AI-assisted Sents \\ \midrule
      BASE  & 215  & 14.2 & - \\
      \ilm  & 170 & 11.4 & 6.6 \\
      \oursys   & 198 & 12.8 & 8.0\\
    \bottomrule
    \end{tabular}
    }
    \caption{Average \# of tokens, sentences, and sentences assisted by AI systems (per response).}
    \label{tab:user-study-stats}
\end{table}

\subsection{Survey feedback}
After each session, we also collect feedback from subjects through a post-session survey. 
The first part of our survey asks participants to recall their experience with \oursys\ mode 
and evaluate various aspects (Table \ref{tab:post-survey-general}). We refer readers to Appendix \ref{appendix:post-survey-rating} for more detailed description. Participants are mildly satisfied with the model performance (3.4 / 5) and are interested in using the system for future WritingPrompt tasks (3.6 / 5), but they are polarized on how easy the system is to learn (3.6 / 5 with a standard deviation of 1.4)

\begin{table}[t]
    \centering
    \scalebox{0.85}{
    \begin{tabular}{lcccc}
    \toprule
      Model   &  Generate & Add & Gen/Sent. & Add/Sent. \\ \midrule
      \ilm & 13.6 & 6.3 & 1.21 & 0.56\\
      \oursys & 16.3 & 7.4 & 1.42 & 0.62\\
    \bottomrule
    \end{tabular}
    }
    \caption{We log every time a subject clicks the \genbutton\ and \addbutton\ buttons, averaged by response and by \# of total sentences.}
    \label{tab:user-study-click}
\end{table} 

\begin{table}[t]
    \centering
    \scalebox{0.7}{
    \begin{tabular}{lccclccc}
    \toprule
       \textbf{Tag}  & \textbf{Pre.} & \textbf{Rec.} & \textbf{F1} & \textbf{Tag}  & \textbf{Pre.} & \textbf{Rec.} & \textbf{F1} \\ 
       \cmidrule(lr){1-4} \cmidrule(lr){5-8}
          \para & 93 & 92 & 92 & \descrip & 89 & 89 & 89 \\
        \bio & 100 & 100 & 100 & \effect & 93 & 93 & 93 \\
        \cause & 81 & 79 & 79 & \idiom & 95 & 90 & 92 \\
        \comparison & 80 & 84 & 82 & \mask & 80 & 81  & 80 \\ \midrule
        \mask(\ilm) & 69 & 68 & 68 & - & - & - & - \\
    \bottomrule
    \end{tabular}
    }
    \caption{Unigram precision, recall, and F1 of model output in comparison with the final text participant submitted. Smaller precision/recall indicates more deletion/insertion operations of participants.}
    \label{tab:user-study-edit}
\end{table}

We also ask them to choose the writing mode they prefer the most and explain their preference. Out of 12 participants, seven prefer \oursys\ writing mode, four prefer \ilm, and only one prefers writing from scratch. The majority of participants favoring the experience of either \ilm\ or \oursys\ demonstrates the potential of AI-assisted writing, especially for open-ended creative tasks like story writing. The most common reason that users prefer the intent-guided generation of \method\ is because it provides fine-grained control over the generated output. The four participants who prefer \ilm\ remark that the system is much simpler to use because it has only one tag (\textit{<mask>}). As one participant comments :  ``\textit{Once I became more comfortable with the remainder of the tags, I think it would be easier for me to write, and therefore more enjoyable. So short-term I would enjoy ILM and then long-term \oursys. As someone who struggled with figuring out what to write next for short stories in elementary school, I wish this existed then!}". 

In the final portion of the survey, we ask them to rate the quality of each tag in \method. If they did not use a certain tag in their writing, they are asked to give a rating for it by recalling their experience of using that tag during the tutorial mode. Table \ref{tab:post-survey-tag} shows that while participants have polarized view about the \idiom\ tag, they are overall satisfied with the output of \para\ and \descrip.

\begin{table}[ht!]
    \centering
    \scalebox{0.85}{
    \begin{tabular}{lcclcc}
   \toprule
    \textbf{Tag} & \textbf{Avg.} & \textbf{Std.} & \textbf{Tag} & \textbf{Avg.} & \textbf{Std.}\\\cmidrule(lr){1-3} \cmidrule(lr){4-6}
         \para & 3.7 & 0.6 & \descrip & 4.0 & 0.8 \\
        \bio & 3.7 & 1.0 & \effect & 3.1 & 0.8\\
        \cause & 3.2 & 0.8 & \idiom & 3.1 & 1.4\\
        \comparison & 3.3 & 1.2 & \mask & 3.6 & 1.3\\
    \bottomrule
    \end{tabular}
    }
    \caption{Post-survey tag-specific ratings on a 5-point Likert scale (1 is negative, 5 is positive).}
    \label{tab:post-survey-tag}
\end{table}


\section{Limitations}

Although our user study demonstrates that  subjects prefer IGA over competing models, it has many limitations. First, NLP researchers are not the right group to ideate the set of writing intents, and in the user study, computer science graduate students are not representative enough as the target users. A more ideal setup is to conduct both the ideation of intents and user study with expert users, preferably English students or teachers. This sort of study could be done on platforms like Upwork. To validate the usefulness of the existing intents in IGA, we also need to conduct interviews with writing professionals and inquire about new prospective intents for future development.
\section{Conclusion}

In this paper, we introduce a new approach to interactive human-AI co-authoring by means of an Intent-Guided Authoring Assistant (\method). Our model is able to infill around author-provided keywords, sentence fragments, and rhetorical instructions with fluent and coherent text. We conduct a small-scale user study which shows that our method has advantages over baseline methods on a creative writing task.
\section*{Ethics statement}

Our data collection is for research purposes only, and thus consistent with the terms of use of all source corpora we mined. For the evaluation process, we strive to compensate the Mechanical Turk workers as well as participants of our user study with competitive payments.

The intended use of \oursys\ is for creative writing. Although generating factually-correct output is not a major focus of creative writing tasks, \method\ often hallucinates facts about real-world entities, a phenomenon that raises ethical concerns and has become an increasing focus in text generation research~\citep{maynez-etal-2020-faithfulness,wang-sennrich-2020-exposure}.
The model can on rare occasions produce offensive outputs, due in large part to GPT-2's pretraining corpora. One potential way to reduce the toxicity of output is to apply profanity filter as a post-processing step before final output is returned.


\section*{Acknowledgements}

We thank the reviewers for the thoughtful comments. We thank Andrew Drozdov, Katherine Thai, Nicholas Monath and other UMass computer science graduate students for helping us with the user study. We thank UMass NLP group for the great advice on the initial draft of this paper. MI was partially supported by award IIS-1955567 from the National Science Foundation (NSF).


\bibliography{anthology,custom}
\bibliographystyle{acl_natbib}

\appendix
\newpage
\section*{Appendix}

\section{Mechanical Turk experiment} \label{appendix:mturk}

We randomly select 50 examples from test set of each tag and get output from \ilm\ and \oursys\ respectively. Each example includes the gold reference and the model output. Each example was assigned to three Mechanical Turk workers who have approval rate higher than 96\% and number of approved HITS greater than 1000. 
Each worker was asked to rate the fluency (FL), coherence (CH) and the adequacy (ADQ) of the infilled content. The first two dimensions are common in natural language generation evaluation, which judge the grammaticality and how well the system output fits into the provided context~\cite{elikyilmaz2020EvaluationOT}.
The last quality dimension ADQ measures how well the infilled content alone fulfill the target author intent. The rating is on a 5-point Likert scale. To increase inter annotator agreement, we collapsed 1 and 2 to 1, 4 and 5 to 3 and change 3 to 2, thus the reported value in \ref{tab:mtruk-res} is reported on a 3-point scale. 

\begin{table}[h]
\scalebox{0.83}{
\begin{tabular}{@{}lccc|ccc@{}}
\toprule
& \multicolumn{3}{c|}{ILM}                       & \multicolumn{3}{c}{\oursys}                        \\ \midrule
& FL            & CH            & ADQ           & FL            & CH            & ADQ           \\\hline
\bio         & $1.97$          & $1.91$          & $1.99$          & $2.13$  & $1.99$  & $2.05$\\
\cause       & $1.82$          & $1.88$        & $1.67^*$        & $2.00$  & $1.99$  & $1.99$\\
\effect      & $1.83$          & $1.73^*$          & $1.79^*$          & $1.94$  & $1.88^*$  & $1.96^*$\\
\comparison  & $1.83$          & $1.79$          & $1.82^*$          & $1.97$  & $1.87$  & $1.94^*$\\
\descrip &   $2.06$            & $2.04^*$          & $2.07^*$          & $2.07$  & $1.95^*$          & $2.06^*$          \\
\idiom       & $1.93^*$          & $1.81^*$          & $1.75^*$          & $1.88^*$          & $1.84^*$& $1.84^*$\\ \bottomrule
\end{tabular}
}
\caption{Ratings of intrinsic crowdsourced evaluation. We collapse the 5-point Likert scale to 3-point scale with 1 (prefer reference), 2 (no preference), 3 (prefer generated text). Fleiss $\kappa$ greater than 0.2 is marked with $^*$. }
\label{tab:mtruk-res}
\end{table}

In general, we find it's hard to get high agreement from the Turkers in terms of fluency except for \idiom\ . Annotators believe the \ilm\ has better fluency mostly because some spans are infilled with clauses rather than short idioms, which leads raters to give higher fluency scores.

\section{Discourse markers used for data extraction} \label{appendix:discourse-markers}
We display discourse markers used for extracting fine-tuning example in Table \ref{tab:cntra-dm}.
\begin{table}[]
    \centering
    \scalebox{0.57}{
    \begin{tabular}{cccc}
    \toprule
    \multicolumn{4}{c}{\comparison}\\ \midrule
         albeit & despite & even though & yet\\
admittedly & even as & even when & by comparison\\
although & even after & even while & by contrast\\
but & even before & even with & conversely\\
but then & even if & however & in contrast\\
but still & even so & in any case & on the contrary\\
concede that & even then & in any event & on the other hand\\
regardless & nevertheless & in spite of & \\
still & nonetheless & whatever & \\
though & no matter & whether & \\ \midrule
\multicolumn{4}{c}{\cause} \\ \midrule
because & as a consequence of & insofar as & on grounds of \\
because of & owing to & not only because of & by dint of \\
due to & by reason of & on account of & thanks to \\
on the strength of & in the wake of & as a result of & by virtue of \\ \midrule
\multicolumn{4}{c}{\effect} \\ \midrule
as a result & accordingly & on that account & thereby \\
consequently & for this reason & inevitably & therefore \\
as a consequence & for that reason & hence & thus \\
that being so & on this account & in the end &  \\
\bottomrule
    \end{tabular}
    }
    \caption{Example discourse markers used for mining fine-tuning example.}
    \label{tab:cntra-dm}
\end{table}

\section{Post-survey rating} \label{appendix:post-survey-rating}

\begin{table}[t]
    \centering
    
    \scalebox{0.83}{
    \begin{tabular}{lcc|lcc}
    \hline
       \textbf{Dimension}  &  \textbf{Avg.} & \textbf{Std.}  & \textbf{Dimension} & \textbf{Avg.} & \textbf{Std.}\\
       \hline
       Fluency & 3.8 & 0.4 & Quality & 2.9 & 0.8\\
        Relevance & 3.6 & 0.8 & Satisfy & 3.4 & 0.8\\
        Coherence & 3.3 & 0.9 & Use Again & 3.6 & 1.0\\
        Interesting & 3.1 & 1.0 & Easy to learn & 3.6 & 1.4\\
        Inspiration & 3.5 & 0.8 & - & - & - \\
    \hline
    \end{tabular}
    }
    \caption{Post-survey general ratings. Ratings are on 5-point Likert scale with 5 being positive experience, 3 neutral, and 1 negative.}
    \label{tab:post-survey-general}
\end{table}

The first section of our survey asks participants to recall their experience with \oursys\ mode and evaluate various aspects presented in Table \ref{tab:post-survey-general}. Besides commonly asked dimensions, such as \textit{fluency}, \textit{relevance}, \textit{coherence}, and general \textit{quality} of system output, we also ask them how often the system generate output that's interesting (\textit{interesting}) and that inspires them to write (\textit{Inspiration}). They are also asked to rate whether they are satisfied with the system output (\textit{satisfy}), whether they would like to use the system again (\textit{Use again}) for the WritingPrompt task, and how easy it is to learn the system (\textit{Easy to learn}). In general, participants are mildly satisfied with the model performance, but understandably, have polarized views on how easy it is to learn this system with standard deviation of 1.4.

\section{Fine-tuning example}

\begin{table*}[]
    \centering
   \scalebox{0.8}{
    \begin{tabular}{p{0.1\linewidth} p{1.0\linewidth}}
    \hline
       Intent Tag  &  Example \\\hline
       \para & \textcolor{blue}{<sub>} the growth potential has consistently declined in this period . \textcolor{blue}{<sub>} \textcolor{blue}{<sep>} The growth potential has been steadily declining throughout this period . \textcolor{blue}{<answer>}\\\hline

\bio & Roger Stone , a Republican strategist , said , `` Issues that were extremely successful for us in the 80 's are n't on the radar screen anymore . '' But Robert Teeter , \textcolor{blue}{<biography>} , insists that the frictions and tensions are simply the growing pains of a governing coalition . \textcolor{blue}{<sep>} the Republican polltaker \textcolor{blue}{<answer>}  \\\hline

\cause & I gawped in astonishment . This morning I read that the University of Exeter has had to employ social media operators to deal with inquiries , \textcolor{blue}{<cause>} increasing \textcolor{blue}{<cause>} email , considering it too slow and unwieldy . \textcolor{blue}{<sep>} because \textcolor{blue}{<answer>} numbers of students will not use \textcolor{blue}{<answer>} \\\hline

\effect & `` I view military prisons as the overlooked campaign of 1864 ; prisons , their management and questions of exchange are taking up a massive part of the bureaucratic part of the war . '' \textcolor{blue}{<effect>} Civil War \textcolor{blue}{<effect>} . \textcolor{blue}{<sep>} In the end , most \textcolor{blue}{<answer>} POWs survived \textcolor{blue}{<answer>} \\\hline

\comparison & Part of being able to extend the network effect of your status update is having the right desktop client for broadcasting updates as well as keeping a lookout on relevant updates from other users . \textcolor{blue}{<concession>} perfect \textcolor{blue}{<concession>} user , we highly recommend the new Seesmic Desktop for managing multiple accounts and tracking custom search results . \textcolor{blue}{<sep>} Though we believe the \textcolor{blue}{<answer>} desktop client is unique to each \textcolor{blue}{<answer>} \\\hline

\descrip & It's because, contrary to what we've been told by satirists, sneering cynics and other such detritus, he is in fact a deeply witty and humane man. \textcolor{blue}{<description>} and he looks like a chimp . \textcolor{blue}{<sep>} He 's intelligent , perceptive \textcolor{blue}{<answer>}  \\\hline

\idiom & As the Senators prepare to face the Montreal Canadiens in Game 3 of their playoff series Sunday night ( CBC , 7 p.m . ET ) at Scotiabank Place , the Ottawa coach had his audience of assembled media \textcolor{blue}{<idiom>} as he tried to deflect any talk about a war of words . \textcolor{blue}{<sep>} in stitches \textcolor{blue}{<answer>} \\\hline

    \end{tabular}
   }
    \caption{Example of each writing intent tag}
    \label{tab:data_examples}
\end{table*}

We display fine-tuning example of each tag (intent) in Table \ref{tab:data_examples}.

\end{document}